\documentclass[a4paper, 10pt, conference]{ieeeconf}      % Use this line for a4
\usepackage{FG2025}
\usepackage{amsmath}
\usepackage{amsfonts}
\usepackage{cite}
\FGfinalcopy % *** Uncomment this line for the final submission

\IEEEoverridecommandlockouts                              % This command is only
\usepackage{graphicx} % for pdf, bitmapped graphics files
\usepackage{multirow}

\def\FGPaperID{****} % *** Enter the FG2025 Paper ID here

\title{\LARGE \bf
Multi-Domain Biometric Recognition using Body Embeddings
}

\author{\parbox{16cm}{\centering
    {\large Anirudh Nanduri*$^1$, Siyuan Huang*$^2$ and Rama Chellappa$^2$}\\
    {\normalsize
    $^1$ University of Maryland, College Park, MD\\
    $^2$ Johns Hopkins University, Baltimore, MD}}
    \thanks{*Equal contribution from the authors.}% <-this % stops a space
}

\begin{document}

\ifFGfinal
\thispagestyle{empty}
\pagestyle{empty}
\else
\author{Anonymous FG2025 submission\\ Paper ID \FGPaperID \\}
\pagestyle{plain}
\fi
\maketitle

%%%%%%%%%%%%%%%%%%%%%%%%%%%%%%%%%%%%%%%%%%%%%%%%%%%%%%%%%%%%%%%%%%%%%%%%%%%%%%%%
\begin{abstract}

Biometric recognition becomes increasingly challenging as we move away from the visible spectrum to infrared imagery, where domain discrepancies significantly impact identification performance. In this paper, we show that body embeddings perform better than face embeddings for cross-spectral person identification in medium-wave infrared (MWIR) and long-wave infrared (LWIR) domains. Due to the lack of multi-domain datasets, previous research on cross-spectral body identification - also known as Visible-Infrared Person Re-Identification (VI-ReID) - has primarily focused on individual infrared bands, such as near-infrared (NIR) or LWIR, separately. We address the multi-domain body recognition problem using the IARPA Janus Benchmark Multi-Domain Face (IJB-MDF) dataset, which enables matching of short-wave infrared (SWIR), MWIR, and LWIR images against RGB (VIS) images. We leverage a vision transformer architecture to establish benchmark results on the IJB-MDF dataset and, through extensive experiments, provide valuable insights into the interrelation of infrared domains, the adaptability of VIS-pretrained models, the role of local semantic features in body-embeddings, and effective training strategies for small datasets. Additionally, we show that finetuning a body model, pretrained exclusively on VIS data, with a simple combination of cross-entropy and triplet losses achieves state-of-the-art mAP scores on the LLCM dataset. %We also demonstrate that while VIS-pretrained face recognition models struggle to perform well on MWIR and LWIR domains with finetuning, a body recognition model achieves significantly better results.

%Clothing changes are typically a major challenge for body identification in the visible domain, but this issue is less pronounced in the infrared domain. We also show that as we move farther away from the visible spectrum, there is less information in face imagery compared to the whole body, making body embeddings to be more effective for identifying subjects. We demonstrate the effectiveness of body-based embeddings on the challenging IARPA Janus Benchmark Multi-Domain Face (IJB-MDF) Dataset.

\end{abstract}

%%%%%%%%%%%%%%%%%%%%%%%%%%%%%%%%%%%%%%%%%%%%%%%%%%%%%%%%%%%%%%%%%%%%%%%%%%%%%%%%
\section{INTRODUCTION}

% Introduce biometric recognition in a general sense that covers both body and face. Then talk about the need for infrared or cross-spectral biometric recognition.
Biometric recognition plays a critical role in a wide range of real-world applications, such as surveillance, authentication, forensic analysis and smart city transportation systems. Biometric modalities can be broadly classified into physiological and behavioral categories. Physiological biometrics include face, fingerprint, body, iris, palmprint etc. Behavioral biometrics aim at capturing unique patterns of human behavior such as gait, voice, hand-writing etc. Advancements in deep learning have significantly enhanced the performance of biometric systems in real-world applications. Face and fingerprint recognition has become the standard for unlocking smartphones. Body recognition (or person re-identification) also has a broad range of applications like smart retail and smart cities apart from security/surveillance. 

Although image- or video-based biometric recognition boasts excellent performance in the visible spectrum even under unconstrained conditions with variations in pose, illumination, and resolution, it becomes a very challenging problem in non-visible spectra such as infrared. Cross-spectral biometric recognition deals with matching RGB (or VIS) data against data acquired beyond the visible spectrum. The infrared spectrum is typically divided into near-, short-, medium-, and long-wave infrared - abbreviated as NIR, SWIR, MWIR and LWIR respectively. NIR and SWIR being the closest to the visible spectrum, share many characteristics of the visible spectrum. For example, similar to visible images, NIR and SWIR images are still predominantly formed from light reflected by the objects, as opposed to MWIR and LWIR which mostly capture heat emitted by the objects. SWIR can capture objects in low-light conditions as well as through fog, smoke or clouds with good resolution. MWIR can work even in conditions of complete darkness since it captures the thermal radiation or the heat signature. LWIR, at the cost of image resolution, can better penetrate atmospheric conditions like fog/smoke compared to MWIR and can also operate in complete darkness.    % Mention some wavelengths

\begin{figure}[h]
	\centering
	\includegraphics[width=0.40\textwidth]{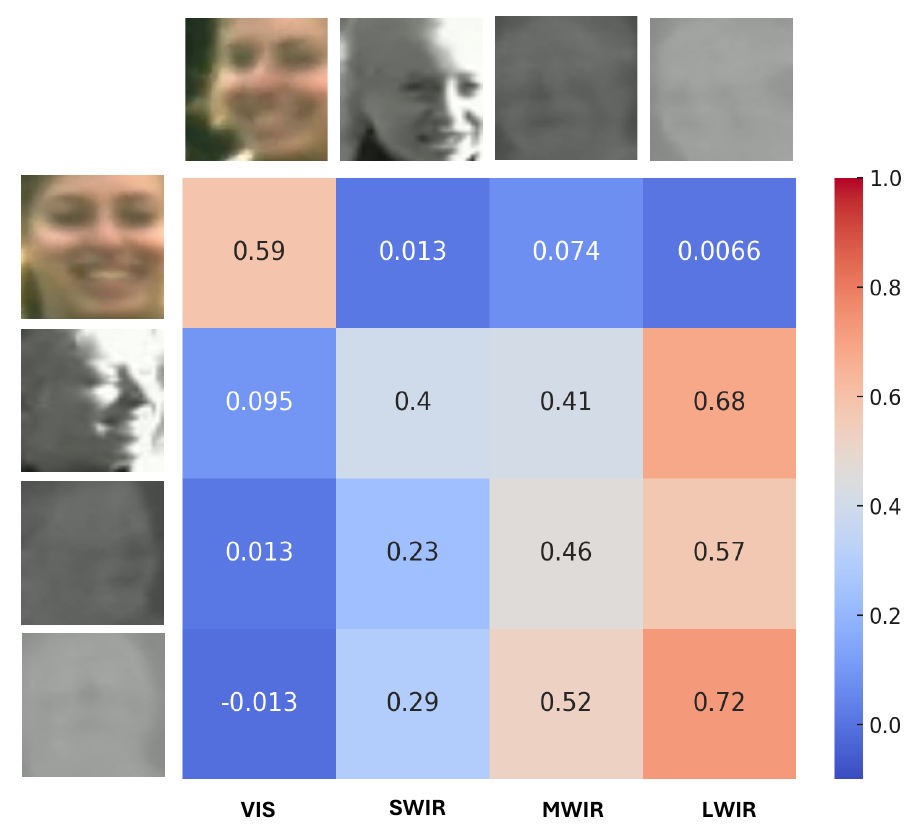}
	\caption{Sample face images from the IJB-MDF dataset from VIS, SWIR, MWIR and LWIR domains along with their cosine similarity score heatmap. Features are extracted using a VIS-pretrained face model.} 
	\label{fig:face}
\end{figure}

\begin{figure}[h]
	\centering
	\includegraphics[width=0.45\textwidth]{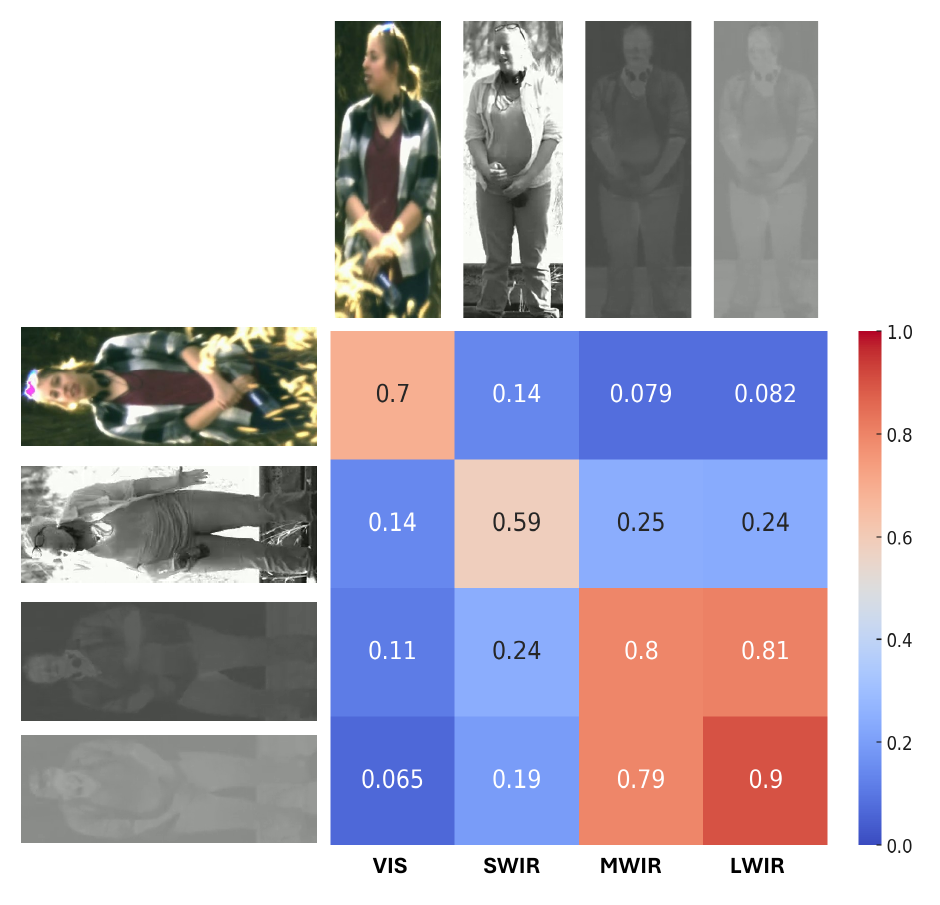}
	\caption{Sample body images from the IJB-MDF dataset from VIS, SWIR, MWIR and LWIR domains along with their cosine similarity score heatmap. Features are extracted using a VIS-pretrained body model.} 
	\label{fig:body}
\end{figure}

Fig. \ref{fig:face} and \ref{fig:body} present sample face and body images from the IJB-MDF \cite{kalka2019iarpa} dataset across VIS, SWIR, MWIR and LWIR domains. The VIS images, captured from a distance of 500m, exhibit significant blurring due to atmospheric turbulence. Although the inrared images were taken from a closer distance of 30m, the MWIR and LWIR images suffer from extremely poor resolution. The facial contours in the MWIR and LWIR images are barely distinguishable, making recognition particularly challenging. In contrast, body outlines remain somewhat more discernable in the MWIR and LWIR body images. The cosine similarity score heatmaps were generated using features extracted from face and body models that were pretrained on VIS datasets. As expected, both models struggle to match VIS to IR domains, as they were trained exclusively on the VIS domain. However, an interesting observation is that MWIR to LWIR matching yields high similarity scores, particularly with body embeddings, suggesting stronger feature consistency across these infrared domains.  

% Introduce face recognition in the context of visible and multi-domain/thermal. And the challenges and advantages of both domains.

% Then compare face and body recognition and the advantages and disadvantages when compared to one another - could talk about availability of datasets, which scenarios either of them are a better choice for, and finally how face is extremely challenging in thermal domains - with example images and cosine-distance-scores for both faces and bodies. 

% Talk about why the scenario we are tackling in this paper is unique and interesting. Also briefly touch upon the ijb-mdf dataset and all the possible problems/protocols that can be studied.

% Talk briefly about our methodology (how we are planning to tackle) and how it is unique compared to other exisiting methods.

Body recognition is commonly referred to as person re-identification (Re-ID) when it focuses on matching pedestrians across non-overlapping camera networks \cite{gu2022clothes,zheng2015scalable}. While visible-light body recognition systems have seen significant advancements, real-world scenarios often present challenging lighting conditions and complex environments where visible cameras may be ineffective \cite{huempowering}. This limitation has led to increasing interest in cross-spectral body recognition, particularly visible-infrared person re-identification (VI-ReID), which requires robust person matching across different spectral domains.

Cross-spectral body recognition introduces unique challenges compared to traditional visible-light body recognition. In addition to the substantial modality gap between visible (RGB) and infrared images, several factors such as loss of color and texture information, low resolution and contrast, and a higher sensitivity to pose variations \cite{wang2022pose}, viewpoint variations \cite{zhu2020aware} and occlusions \cite{zhang2022fine} need to be addressed. Moreover, thermal patterns can fluctuate significantly due to changes in physical activity or environmental conditions, while visible appearances tend to remain relatively stable. Recent progress in deep learning has led to various methods that address these challenges. Early approaches focused on reducing the modality gap through feature-level alignment \cite{wang2019learning}, utilizing techniques such as adversarial learning to minimize the distribution discrepancy between visible and infrared feature spaces. Subsequent research introduced dual-stream architectures \cite{huempowering,wu2024wrim,ren2024implicit} that process visible and infrared images separately before performing cross-modal feature fusion. These methods often incorporate modality-specific feature extraction modules to preserve spectrum-unique features while maintaining identity-discriminative information.

Transformers and attention mechanisms \cite{vaswani2017attention} have introduced significant advancements in cross-spectral body recognition. These methods leverage self-attention to capture long-range dependencies and cross-attention to align features across modalities \cite{feng2022visible,di2019polarimetric}. Some approaches also explored knowledge distillation strategies to transfer discriminative features learned from the visible domain to guide the learning of infrared features \cite{zhou2022knowledge}. 

%Beyond the modality gap, cross-spectral body recognition must also address common challenges in person body recognition, including variations in pose \cite{wang2022pose}, viewpoint \cite{zhu2020aware}, and occlusion \cite{zhang2022fine}. The problem becomes particularly challenging when combining these variations with the inherent modality differences. For instance, thermal patterns can vary significantly with changes in physical activity or environmental conditions, while visible appearances may remain relatively stable.

While most existing approaches limit their scope to a single infrared band, we introduce the novel challenge of multi-domain body recognition spanning SWIR, MWIR, and LWIR domains. Unlike conventional methods that specialize in a specific spectrum, our approach requires a single model to deliver consistent performance across VIS-to-SWIR, VIS-to-MWIR, and VIS-to-LWIR recognition, effectively bridging the gap between multiple spectral domains.
%Most existing methods focus on bridging the modality gap through feature alignment or learning shared representations. However, these methods may not fully exploit the complementary nature of visible and infrared information. Thermal patterns captured by infrared cameras could provide additional biometric cues that are not available in visible images, potentially enabling more robust identification under challenging conditions. This suggests the need for methods that can effectively leverage the unique advantages of each modality while maintaining robust cross-spectral matching capabilities.

In this paper, we investigate the efficacy of body embeddings for cross-domain biometric identification across multiple domains where face embeddings struggle. We perform experiments on the IARPA Multi-Domain Face dataset (IJB-MDF) \cite{kalka2019iarpa} which comprises images and videos captured using a variety of cameras: fixed and body-worn, capable of imaging at visible, short-wave, mid-wave, and long-wave infrared wavelengths at distances up to 500m. We introduce a new benchmark protocol to evaluate cross-spectral body identification performance for VIS-to-SWIR, VIS-to-MWIR, and VIS-to-LWIR settings. We leverage a vision transformer model \cite{huang2023self} that extracts local semantic features along with global features to extract body embeddings. % we use the AdaFace \cite{kim2022adaface} model to obtain face embeddings.

We also show that a simple finetuning strategy can achieve state-of-the-art mAP scores and competitive rank-1 accuracies on LLCM \cite{zhang2023diverse}, the latest VI-ReID benchmark dataset, on both visible-to-infrared and infrared-to-visible modes.

%Recently, this problem has been tackled through novel network architectures involving attention mechanisms that aim to extract multi-scale embeddings that are domain invariant \cite{wu2024wrim}. 

%Conclude with the contributions of the paper + sections outline

The main contributions of this paper are:
\begin{itemize}

\item We formulate the problem of multi-domain cross-spectral body recognition, incorporating domains like short-wave (SWIR), medium-wave (MWIR), and long-wave infrared (LWIR), and establish benchmark results on the IJB-MDF dataset.
\item We compare the performance of face-embeddings and body-embeddings across SWIR, MWIR and LWIR domains on the IJB-MDF dataset, highlighting the advantages of body-based recognition in cross-spectral settings.
\item We demonstrate that a body model pretrained on a VIS dataset can achieve state-of-the-art mAP scores on the LLCM dataset with simple fine-tuning, without the need for complex domain adaptation techniques.
\item We investigate the significance of the local semantic features in body embeddings as we move away from the visible spectrum towards long-wave infrared, revealing the impact of different body regions across infrared modalities.
\item We provide insights into domain gap between VIS, SWIR, MWIR and LWIR, through a comparative analysis of the model performance when trained on various domain combinations, enhancing understanding of cross-spectral learning.

\end{itemize}

%%%%%%%%%%%%%%%%%%%%%%%%%%%%%%%%%%%%%%%%%%%%%%%%%%%%%%%%%%%%%%%%%%%%%%%%%%%%%%%%
\section{Related Work}

Biometric recognition in the visible domain has been well-studied using diverse modalities such as face, body, fingerprint, iris, palmprint etc. Face recognition performance has seen considerable improvement since the introduction of margin-based loss functions \cite{liu2017sphereface,deng2019arcface,wang2018cosface,kim2022adaface}. 
Recent body recognition works have explored various strategies to handle appearance variations. Gu et al. \cite{gu2022clothes} developed a clothing-invariant feature learning method using adversarial training, while Yang et al. \cite{yang2023good} proposed a causality-based framework to mitigate appearance bias. Han et al. \cite{han2023clothing} introduced feature augmentation methods to address clothing changes.

%Although these models maintain their performance on non-visible domains like SWIR, they fail to generalize on the more challenging domains like MWIR and LWIR owing to the larger domain discrepancy. 
% VIS Person body recognition - A brief paragraph 

%Cross-spectral biometric recognition: Cross-spectral Face Recognition and VIbody recognition
%Cross-spectral biometric recognition is more challenging owing to the domain discrepancy between the visible and non-visible spectra, and the approaches to solve this problem can be largely divided into two categories: (i) feature-level methods that learn a shared feature representation and image-level methods that learn to transform images from one domain to another.

Cross-spectral biometric recognition methods can be broadly categorized into image-level and feature-level approaches. The image-level approaches aim to bridge the domain gap by transforming all images into a common domain before extracting their features. Feature-level methods, on the other hand, focus on learning a shared feature representation that remains consistent across different domains.

% Face Cross-spectral related works - touch briefly
\subsection{Cross-spectral face recognition:} He et al. \cite{he2021coupled} proposed Coupled Adversarial Learning which combines image- and feature-level approaches. Luo et al. \cite{luo2022memory} proposed Memory-Modulated Transformer Network, which is an image-level approach. Yang et al. \cite{yang2023robust} proposed a feature-level approach called the RPC network which involves estimating pseudo-labels and contrastive learning to obtain domain-independent representations. The methods described so far deal with NIR-VIS face recognition. Due to the larger domain-gap of the MWIR and LWIR domains, image-level approaches based on GAN architectures \cite{zhang2018tvgan,chen2019sggan,iranmanesh2020coupled,peri2021synthesis,chen2022attention,anghelone2023anyres} are more popular.

\subsection{Cross-spectral body recognition:} %has gained increasing attention due to its practical importance in 24-hour smart city systems. 
While VI-ReID research has primarily focused on feature-level methods that learn modality-shared representations, some image-level approaches have also been explored. In particular, GAN-based methods such as ThermalGAN \cite{kniaz2018thermalgan} and AlignGAN \cite{wang2019rgb} aim to bridge the modality gap by synthesizing cross-spectral images, enabling better alignment between visible and infrared domains.

 Recent methods focus on extracting modality-invariant representations by disentangling them from modality-specific information through novel network modules and loss functions. Feng et al. \cite{feng2022visible} introduced the Cross-Modality Interaction Transformer (CMIT), which enhances discriminative feature learning by leveraging CLS token interactions between visible and infrared modalities and employing a modality-discriminative loss function to improve feature alignment. Zhang et al. \cite{zhang2023diverse} proposed DEEN that generates diverse embeddings using a Diverse Embedding Expansion module to learn more informative feature representations that reduce the domain gap. Wu et al. \cite{wu2024wrim} developed WRIM-Net that mines modality-invariant information using a Multi-dimension Interactive Informative Mining (MIIM) module and an Auxiliary-Information-based Contrastive Learning (AICL) approach. Ren et al. \cite{ren2024implicit} introduced IDKL, a model that uses implicit discriminative knowledge within modality-specific features to improve the discriminative power of modality-shared features. Alehdaghi et al. \cite{alehdaghi2025cross} proposed the Mixed Modality Erased and Related (MixER) method, which disentangles modality-specific and modality-shared identity information using a combination of orthogonal loss, modality-erased identity loss, modality-aware loss, and mixed cross-modal triplet loss functions. Additionally, they introduce a new mixed-modal ReID setting, further expanding the scope of cross-spectral person re-identification.%Hu et al. \cite{huempowering} proposed TVI-LFM that uses large foundation models to generate text descriptions that improve infrared representations in cross-spectral body recognition by addressing the lack of color information.

\textbf{Local Feature Learning.} The importance of local features has been recognized in both visible and cross-spectral body recognition. Zhang et al. \cite{zhang2022fine} developed a multi-feature fusion network for global and partial feature extraction. Zhang et al. \cite{zhang2022person} introduced a hierarchical discriminative spatial aggregation method that focuses on local human parts to handle spatial misalignment. In the cross-spectral domain, Wu et al. \cite{wu2022discriminative} proposed the Discriminative Local Representation Learning (DLRL) model, designed to capture robust fine-grained feature representations. 

%\textbf{Knowledge Distillation in Cross-spectral Body Recognition.} Several works have explored knowledge distillation to transfer discriminative features across modalities. Wang et al. \cite{wang2019learning} proposed a dual-teacher distillation framework to guide the learning of infrared features using visible domain knowledge. Shi et al. \cite{shi2024learning} proposed a method based on progressive contrastive learning with hard and dynamic prototypes to address the limitations of using only cluster centers by incorporating divergence and variety. However, the effectiveness of knowledge distillation is limited by the significant domain gap and the lack of paired cross-spectral training data.

\textbf{Multimodal ReID.}
Li et al. \cite{li2024all} introduced an All-in-One framework for multimodal ReID in the wild, incorporating diverse modalities such as VIS, infrared, sketches, and text. Similarly, He et al. \cite{he2024instruct} proposed a multi-purpose ReID model capable of handling six distinct ReID tasks, including clothes-changing ReID, VI-ReID, and text-to-image ReID.

While these approaches are categorized as multimodal ReID and treat VI-ReID as a single task, we address the challenge of multimodal VI-ReID (or multi-domain cross-spectral body recognition) by explicitly considering multiple infrared bands—SWIR, MWIR, and LWIR. In this task, a single model must effectively operate across multiple IR bands alongside VIS, thereby pushing the field beyond traditional single-modality cross-spectral approaches and enabling more robust person recognition across spectral domains.

%%%%%%%%%%%%%%%%%%%%%%%%%%%%%%%%%%%%%%%%%%%%%%%%%%%%%%%%%%%%%%%%%%%%%%%%%%%%%%%%
\section{Methodology}

\subsection{Cross-Spectral Semantic Body Identification}

\begin{figure*}[h]
	\centering
	\includegraphics[width=0.9\textwidth]{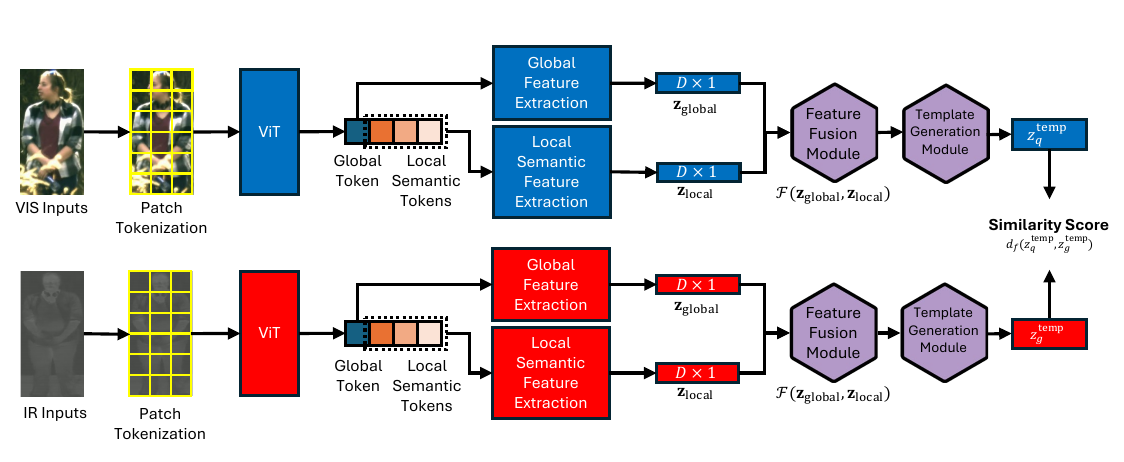}
	\caption{Cross-Spectral Body Identification Pipeline - the VIS and IR inputs are passed through a ViT architecture to generate global and local semantic features, which are fused to generate the final feature. All the fused features corresponding to a single template are then used to generate the template feature. The template feature of the query template is then matched against all the gallery templates to estimate the subject id.} 
	\label{fig:pipeline}
\end{figure*}
   
Fig. ~\ref{fig:pipeline} illustrates our proposed pipeline, which address the significant modality discrepancy between RGB and infrared images through semantic-aware feature learning. The key insight is that while low-level appearance features differ drastically between spectrum, high-level semantic body structures remain consistent and can serve as reliable matching cues.

While local semantic features are crucial for body identification \cite{tang2023humanbench, leng2019survey, zhu2022pass}, extracting consistent semantics across RGB and infrared domains is particularly challenging due to the distinct imaging principles. We leverage a model \cite{huang2023self} that utilizes anatomically-aware keypoint prompts to guide the Interactive Segmentation Model, ensuring consistent semantic extraction across spectrum.

Our model processes both RGB and infrared inputs through parallel streams while maintaining semantic correspondence. For inference, given a query-gallery pair $(x_q, x_g)$ from each spectrum, our model extracts multi-granular features through a feature extraction process. The model first computes global features $\mathbf{z}_q^{\text{global}}$ and $\mathbf{z}_g^{\text{global}}$ to capture overall shape patterns for each spectrum. Simultaneously, it extracts local semantic features ${\mathbf{z}_q^{\text{local}}}$ and ${\mathbf{z}_g^{\text{local}}}$ that encode spectrum-invariant structural information for each semantic area. These features are then combined into concatenated representations through a learnable fusion module
\begin{equation}
\mathbf{z}_q = \mathcal{F}(\mathbf{z}_q^{\text{global}}, \mathbf{z}_q^{\text{local}}), \quad
\mathbf{z}_g = \mathcal{F}(\mathbf{z}_g^{\text{global}}, \mathbf{z}_g^{\text{local}}).
\end{equation}

All the fused query features corresponding to a single query template are then used to generate the final query template feature (typically through averaging) - $\mathbf{z}_q^{\text{temp}}$. Similarly, we generate the gallery template $\mathbf{z}_g^{\text{temp}}$.

During matching, we use a similarity computation
\begin{equation}
s(x_q^{\text{temp}}, x_g^{\text{temp}}) = d_f(\mathbf{z}_q^{\text{temp}}, \mathbf{z}_g^{\text{temp}}),
\end{equation}
where $d_f$ is a distance function that measures feature similarity, and $x_q^{\text{temp}}$, and $x_g^{\text{temp}}$ refer to the set of query and gallery media corresponding to a specific template. This semantic-guided matching strategy effectively bridges the spectrum gap by leveraging shared structures that are consistent across spectrum. The method focuses on high-level semantic patterns rather than low-level appearance details, enabling fine-grained comparison through part-based matching while maintaining robustness to spectrum-specific variations through adaptive feature fusion.

\subsection{Network Architecture and Loss Functions} \label{network}

% ViT architecture
The backbone of our cross-spectral model uses a Vision Transformer (ViT) \cite{dosovitskiy2020image} architecture that divides input images into non-overlapping patches and processes them through a series of transformer blocks. Specifically, given an input image $x \in \mathbb{R}^{H \times W \times C}$, we first partition it into patches $\mathbf{x}_p$ of size $P \times P$, resulting in a sequence of $N = HW/P^2$ patches. These patches are then linearly projected to create patch embeddings
\begin{equation}
\mathbf{E} \leftarrow [\mathbf{x}_{\text{global}}; \mathbf{x}_{\text{local}}; \mathbf{x}_p^1\mathbf{E}; ...; \mathbf{x}_p^N\mathbf{E}] + \mathbf{E}{\text{pos}},
\end{equation}
where $\mathbf{E} \in \mathbb{R}^{(N+1) \times D}$ represents the patch embeddings, $D$ is the embedding dimension, $\mathbf{x}_{\text{global}}$ is the global token, $\mathbf{x}_{\text{local}}$ is the local token consists of different local areas, including face, torso, and lower body, and $\mathbf{E}{\text{pos}}$ is learnable position embeddings.

The semantic features are processed through the ViT encoder before being fused with the global features:
\begin{equation} \label{eq:final_feature}
\mathbf{z}_{\text{final}} = \text{FFN}([\mathbf{z}_{\text{global}}; \mathbf{z}_{\text{local}}])
\end{equation}
where FFN is a feed-forward fusion network that adaptively combines global and local semantic features.

Our training includes multiple objectives to effectively align cross-spectral features while preserving discriminative semantic information. The overall training objective combines identity-based learning with semantic consistency constraints to handle the spectrum discrepancy between RGB and infrared domains.
Given a batch of image features $\mathbf{Z}_{\text{final}}$ generated from the model, We first use an identity loss to enforce identity consistency across spectrum, i.e.,
\begin{equation}
\mathcal{L}_{\text{id}} = -\frac{1}{N}\sum_{i=1}^N \log \frac{\exp(\mathbf{Z}_{\text{final}_i}^{y_i})}{\sum_{j=1}^N \exp(\mathbf{Z}_{\text{final}_i}^{y_j}))},
\end{equation}
where $N$ is the batch size, and $y_i$ represents the identity label of $\mathbf{Z}_{\text{final}_i}$. To further improve the feature discrimination, we include a triplet loss \cite{hermans2017defense} that enforces margin constraints. i.e.,
\begin{equation}
\mathcal{L}_{\text{tri}} = \sum_{i,j=1}^N [\Delta + d(\mathbf{Z}_{\text{final}_i}, \mathbf{Z}_{\text{final}_j}) - \min_{i\neq j} d(\mathbf{Z}_{\text{final}_i}, \mathbf{Z}_{\text{final}_j})]_+,
\end{equation}
where $\Delta$ is the margin parameter, $d(\cdot,\cdot)$ measures the feature distance, and $[\cdot]_+$ denotes the hinge function.

The final training loss combines these losses with balanced weights, i.e.,
\begin{equation}
\mathcal{L} = \mathcal{L}_{\text{id}} + \lambda\mathcal{L}_{\text{tri}},
\end{equation}
where $\lambda$ is hyperparameter controlling the contribution of each loss term. This joint optimization enables the model to learn robust features while preserving identity-discriminative information at both global and local semantic levels. During training, we use a hard mining strategy to select the most informative negative samples, which helps improve the model's discriminative power across spectrum.

\subsection{Face Recognition Model}

We employ the AdaFace \cite{kim2022adaface} model for our face recognition experiments. AdaFace employs a margin-based loss function that prioritizes hard samples with high image quality while de-emphasizing hard samples of low quality, where image quality is estimated based on the feature norm. We select AdaFace for our experiments on challenging infrared (IR) domains due to its remarkable performance, particularly on mixed and low quality datasets. 

As our baseline, we use the ResNet100 model pretrained on the WebFace12M dataset \cite{zhu2021webface260m}. As shown by the cosine similarity scores in Fig. \ref{fig:face}, while AdaFace features exhibit poor VIS-to-IR matching, they maintain identity consistency within the IR domains, demonstrating their effectiveness in handling low quality images.

\section{Experiments and Results}

\subsection{Datasets and Protocols}

% Brief description
The \textbf{IARPA JANUS Benchmark Multi-domain Face (IJB-MDF)} dataset consists of images and videos of 251 subjects captured using a variety of cameras corresponding to visible, short-, mid-, and long-wave infrared and long range surveillance domains. There are 1,757 visible enrollment images,  40,597 short-wave infrared (SWIR) enrollment images and over 800 videos spanning 161 hours.

% The ground-truth information
% Generating the body ground-truth using face and Yolov10 

In this paper, we choose the VIS-500m (visible spectrum videos captured at a distance of 500m) as the gallery domain, and the SWIR-30m, MWIR-30m, and LWIR-30m as the query domains. The dataset provides ground-truth bounding boxes for faces in every frame of the videos. We generate bounding boxes for bodies using the YOLOv10-X \cite{THU-MIGyolov10} model. We then generate the labels for the detected bounding boxes by matching them against the face boxes with the maximum overlap. To avoid mislabeling, we ignore the detections that result in more than 0.75 IoU score with more than one face. There are 31 videos in the VIS-500m domain and 42 videos in each of SWIR-30m, MWIR-30m, and LWIR-30m domains. 
% Explain the face and body protocols - gallery-1, gallery-2: train, test

The dataset divides the 251 subjects into two disjoint galleries: gallery-1 with 126 subjects and gallery-2 with 125 subjects. Since there are no training protocols provided with the dataset, we use the videos corresponding to the gallery-1 subjects as the training data and the gallery-2 subjects are included in the test data. We follow the subject-specific modeling paradigm as in \cite{kalka2019iarpa} and \cite{nanduri2024template}, where all the media corresponding to a subject are combined into a single template. The evaluation protocol is 1:N identification, where each query template is matched against the gallery templates to predict the subject id. 

% Compare with LLCM and SYSU datasets
Compared to the benchmark datasets used for the problem of VI-ReID - RegDB \cite{nguyen2017person}, SYSU \cite{wu2017rgb} and LLCM \cite{zhang2023diverse}, the IJB-MDF dataset offers unique challenges. Firstly, the videos corresponding to the visible spectrum are captured at a much longer distance (500m) compared to the infrared videos (captured at 30m), resulting in a lot of atmospheric turbulence and blur. There are also more occlusions because each video captures multiple persons (walking away from, and toward the camera). Secondly, the training dataset for IJB-MDF consists of only 126 subjects, of which there are only about 63 subjects that have videos from all four domains. In contrast, the SYSU and LLCM datasets have 395 and 713 identities for training, respectively. This makes the training more challenging due to the smaller training data size. And finally, the IJB-MDF dataset distinctly separates data from the SWIR, MWIR and LWIR bands of the infrared spectrum, allowing us to work on the multi-domain body recognition problem. On the other hand, RegDB only has VIS and thermal images, while SYSU and LLCM contain VIS and NIR images. 

\textbf{Low-Light Cross-Modal (LLCM)} is the latest VI-ReID benchmark dataset that contains 46,767 images of 1,064 identities, captured using a 9-camera network in low-light environments. It features VIS and IR images under diverse climate conditions and clothing styles. The train:test split of about 2:1 leads to 30,921 bounding boxes (VIS: 16,946 and IR: 13,975) of 713 identities in the training set and 13,909 bounding boxes(VIS: 13,909 and IR: 7,166) of 351 identities in the test set. The dataset uses both the VIS to IR mode and IR to VIS mode to evaluate the performance of VI-ReID models. 

\subsection{Implementation Details}
%\subsubsection{Face}
%We perform experiments using the pretrained ArcFace, CosFace and AdaFace models. 
% ArcFace, CosFace and AdaFace
% (Finetune without face alignment?)
% NormPooling to filter out occluded and very bad faces - has to improve performance

% (Class-center translation for domain adaptation without training? How can features of a network be translated without losing their discriminative capabilities? Some sort of manifold mapping and de-mapping?)
Our base architecture is the ViT-Base-Patch16 vision transformer. It is pretrained on the LUPerson \cite{fu2020unsupervised} dataset in a self-supervised manner, followed by finetuning on the IJB-MDF dataset. The margin parameter $\Delta$ in the triplet loss is set to 0, and the hyperparameter $\lambda$ in the final loss is set to 1, so that equal weight is given to the triplet and cross-entropy losses. The feature fusion module first averages the local semantic features corresponding to the face, torso and lower body, and then concatenates the global feature and the averaged local feature. The local and global features are of size $1 \times 768$, so the size of the final fused feature is $1 \times 1536$.

The training data consists of cropped and resized images (of size $384 \times 128 $) extracted from each frame of the training videos. The training batches are formed by sampling the data such that each subject in a batch has images from each of the four domains of interest - VIS, SWIR, MWIR and LWIR. This 'domain-aware sampling' results in a more consistent performance across all three infrared domains compared to random sampling. 

During inference and matching, we use the cosine similarity metric to match the query and gallery template features. We generate the template features by taking the average of all the features corresponding to the same id. In all our experiments on the IJB-MDF dataset, we report separate Rank-1 accuracy metrics for VIS-to-SWIR, VIS-to-MWIR, and VIS-to-LWIR matching, providing a detailed performance evaluation of a model across different infrared domains. The LLCM evaluation protocol does not include template generation and we report both the Rank-1 and mAP scores for VIS-to-IR and IR-to-VIS person re-identification scenarios.

% Sembody recognition model - baseline and finetuned
% DEEN model - finetuned on MDF
% Sembody recognition model - part based ablation - without face, without torso, without lower body
% Domain-aware sampling and its advantage with Triplet loss
% Possible fusion results: norm-pooling based fusion to filter out the poor quality data
% Score-level and feature level fusion

\subsection{Results}
% Baseline and finetuned AdaFace 

\textbf{Comparison of Face and Body Embeddings from Pretrained and Finetuned Models.} 
%One of the unique advantages of the IJB-MDF dataset is that it has ground-truth face bounding boxes and also an almost full torso  
We utilize the ResNet-101 architecture with the AdaFace loss \cite{kim2022adaface} to extract face embeddings. We use the checkpoint pretrained on the WebFace-12M \cite{zhu2021webface260m} dataset as our baseline. 
As shown in Table \ref{tab:face_baseline}, the baseline face model performs significantly better than the body models on the easier SWIR domain. However, the AdaFace model struggles to extract meaningful embeddings from the MWIR and LWIR domains that can be effectively matched against VIS embeddings. Even after finetuning on the IJB-MDF dataset, the performance only marginally improves on MWIR and LWIR, while decreasing on the SWIR domain. 

In contrast, while the body model pretrained on the LUPerson \cite{fu2020unsupervised} dataset performs worse on SWIR, it outperforms the face model on MWIR and LWIR domains. Furthermore, after finetuning on IJB-MDF, its performance drastically improves across all three domains.

Due to the significant gap in the performance of face and body embeddings, particularly on MWIR and LWIR domains, neither feature-fusion nor score-based fusion provides any improvement over using body-embeddings alone. 

\begin{table}[t!]
	\renewcommand{\arraystretch}{1.5}
	\caption{\label{tab:face_baseline}Performance of Pretrained Face and Body recognition models before and after finetuning on the IJB-MDF Dataset - Rank-1 (\%)}
	\centering
		\renewcommand{\arraystretch}{1.8}
		\begin{tabular}{c c c c} 
			\hline
			\textbf{Model} &  \textbf{SWIR} & \textbf{MWIR} & \textbf{LWIR} \\ 
			\hline
			%CosFace & 67.19 & 1.56 & 1.56 \\
			%\hline
			%ArcFace & 65.62 & 1.56 & 1.56 \\
            AdaFace (pretrained) & \textbf{87.50} & 0.00 & 3.12 \\
            AdaFace (finetuned) & 59.38 & \textbf{15.62} & \textbf{15.62} \\
            %AdaFace (partially finetuned) & 60.94 & \textbf{17.19} & 12.50 \\
			%\hline
			\hline
            Body Model (pretrained) & 38.10 & 9.52 & 11.11 \\
            Body Model (finetuned) & \textbf{77.78} & \textbf{77.78} & \textbf{77.78} \\
            \hline
	\end{tabular}
\end{table}

\textbf{Comparison with SoTA methods on the LLCM Dataset.}
We validate the performance of our approach by finetuning the pretrained body model on LLCM \cite{zhang2023diverse}. We compare the results of our finetuned model against the latest state-of-the-art models namely - DEEN \cite{zhang2023diverse}, WRIM-Net \cite{wu2024wrim}, IRM or Instruct-ReID \cite{he2024instruct}, IDKL \cite{ren2024implicit}, and MixER \cite{alehdaghi2025cross}. These results are presented in Table \ref{tab:reid_comparison}. 

We follow the evaluation protocol outlined in the publicly released code of LLCM/DEEN. IRM (STL) \cite{he2024instruct} refers to the Single-Task Learning scenario, where the model is trained and tested exclusively on the LLCM dataset. For IDKL, we report the results from \cite{alehdaghi2025cross}, which do not include re-ranking, to ensure a fair comparison with the other results. 

The results indicate that our model outperforms all the other models on the mAP metric in both VIS-to-IR and IR-to-VIS modes. On the VIS-to-IR mode, we achieve a Rank-1 score comparable to the state-of-the-art MixER model. \textit{This is particularly noteworthy given that, unlike state-of-the-art models, which incorporate specialized modules or novel loss functions to mitigate domain discrepancies, our model - pretrained solely on a VIS dataset - relies only on a simple combination of cross-entropy and triplet losses.}

\begin{table}[t!]
\renewcommand{\arraystretch}{1.5}
    \centering
    \caption{Comparison of Rank-1 and mAP scores for VIS-to-IR and IR-to-VIS modes on LLCM Dataset. *All numbers are reported without re-ranking to ensure a fair comparison. }
    \begin{tabular}{l|cc|cc}
        \hline
        \multirow{2}{*}{\textbf{Model}} & \multicolumn{2}{c|}{\textbf{VIS to IR}} & \multicolumn{2}{c}{\textbf{IR to VIS}} \\
        
        & Rank-1 & mAP & Rank-1 & mAP \\
        \hline
        %DDAG \cite{DDAG} & 40.3 & 48.4 & 48.0 & 52.3 \\
        %CAJ \cite{CAJ} & 56.5 & 59.8 & 48.8 & 56.6 \\
        DEEN \cite{zhang2023diverse} & 62.5 & 65.8 & 54.9 & 62.9 \\
        WRIM-Net \cite{wu2024wrim} & 67.0 & 69.2 & 58.4 & 64.8 \\
        IRM (STL) \cite{he2024instruct} & 64.9 & 64.5 & \textbf{66.2} & 66.6 \\
        %IRM (MTL) $\dagger$ \cite{he2024instruct} & 65.7 & 67.2 & \textbf{66.7} & 67.5 \\
        IDKL* \cite{ren2024implicit}  & 70.4 & 55.0 & 62.5 & 49.3 \\
        MixER \cite{alehdaghi2025cross} & \textbf{70.8} & 56.6 & 65.8 & 51.1 \\
        \hline
        Our Model & \textbf{70.5} & \textbf{72.9} & 64.1 & \textbf{70.3} \\
        \hline
    \end{tabular}
    
    \label{tab:reid_comparison}
\end{table}

% DomainAware Sampler and hard-mining statistics?
% Part-based ablation
\textbf{Ablation Study of Local Semantic Features.}
As mentioned in section \ref{network}, we construct the final feature by concatenating the global and local features. The local feature is computed as in (\ref{eq:local_feature}), by averaging the three local semantic features corresponding to the face, torso and lower body regions. 
\begin{equation} \label{eq:local_feature}
\mathbf{z}_{\text{local}} = (\mathbf{z}_{\text{face}} + \mathbf{z}_{\text{torso}} + \mathbf{z}_{\text{lower-body}})/3,
\end{equation}
\begin{equation} \label{eq:final_feature2}
\mathbf{z}_{\text{final}} = [\mathbf{z}_{\text{global}}; \mathbf{z}_{\text{local}}]
\end{equation}

In Table \ref{tab:local_part_ablation}, we analyze the impact of individual local semantic feature across the three IR domains. By comparing the results within each column, we observe the following:
\begin{itemize}
    \item \textbf{SWIR}: Removing face features leads to a slight performance drop, while eliminating torso or lower-body features results in a slight improvement. 
    \item \textbf{MWIR}: Performance drops significantly when either the face or lower-body features are removed, whereas removing the torso has only a minor effect.
    \item \textbf{LWIR}: Eliminating any of the three features causes a drastic decline in performance, highlighting the importance of all body regions in this domain.
\end{itemize}

\begin{table}[t]
	\renewcommand{\arraystretch}{1.2}
	\caption{\label{tab:local_part_ablation}Impact of local semantic features on performance - \\ Rank-1 (\%)}
	\centering
		\renewcommand{\arraystretch}{1.2}
		\begin{tabular}{c c c c } 
			\hline
			\textbf{Final Feature Construction} &  \textbf{SWIR} & \textbf{MWIR} & \textbf{LWIR} \\ 
			\hline
			global + face + torso + lower-body & 77.78 & \textbf{77.78} & \textbf{77.78}\\
			global + torso + lower-body & 76.19 & 73.02 & 69.84 \\
            global + face + lower-body & \textbf{79.37} & 76.19 & 66.67 \\
            global + face + torso & \textbf{79.37} & 74.60 & 73.02 \\
            %global + face & 77.78 & 74.60 & 80.77 \\
			\hline
	\end{tabular}
\end{table}

\textbf{Analysis of Domain Gap.} 
To gain a better understanding of the domain-gap between the VIS and the IR domains, as seen from the eyes of a deep network, we present results using a model that was pretrained solely on a VIS dataset (LUPerson) in Table \ref{tab:dif_gallery}. The Table shows cross-domain results with various pairs of gallery and query domains. The most interesting observation from the results is that, although the network was trained only on VIS data, the cross-domain performance between MWIR-LWIR is extremely high (95.58\%). This shows that the feature spaces of MWIR and LWIR domains are fairly similar even though the model has only seen VIS data during training. And to a lesser degree, the feature spaces of VIS and SWIR domains are also close. This also supports the case for why feature-level approaches can work so well even when the domain-gap is very high - if the feature spaces can be brought together without losing the class-discriminative ability of the features, we can solve the problem of cross-spectral biometric recognition without needing to synthesize images from different domains.

%TODO results with adaface features also

In Table \ref{tab:swir_only} we report results with models trained on different training domain combinations. Notably, training a model using only VIS and SWIR data leads to improvements in both the MWIR and LWIR domains. However, the greater performance gain in MWIR compared to LWIR suggests a stronger overlap of discriminative information between SWIR and MWIR features, as opposed to SWIR and LWIR.

\begin{table}[t!]
	\renewcommand{\arraystretch}{1.5}
	\caption{\label{tab:dif_gallery}Cross-domain performance with different gallery-query combinations on IJB-MDF dataset - Rank-1 (\%)}
	\centering
		\renewcommand{\arraystretch}{1.8}
		\begin{tabular}{c c c c c} 
			\hline
			\textbf{Gallery $\downarrow$ Probe $\rightarrow$ } &  \textbf{VIS}&  \textbf{SWIR}&  \textbf{MWIR}&  \textbf{LWIR} \\ 
			\hline
			%CosFace & 67.19 & 1.56 & 1.56 \\
			%\hline
			%ArcFace & 65.62 & 1.56 & 1.56 \\
            \textbf{VIS} & - & 26.95 & 5.00 & 3.55 \\
            \textbf{SWIR} & 42.55 & - & 12.85 & 10.36 \\
            \textbf{MWIR} & 11.43 & 30.52 & - & 95.58 \\
            \textbf{LWIR} & 11.35 & 27.09 & 95.98 & - \\
			%\hline
			\hline
	\end{tabular}
\end{table}

\begin{table}[t!]
	\renewcommand{\arraystretch}{1.5}
	\caption{\label{tab:swir_only}Cross-spectral Performance with Different Training Domains - Rank-1 (\%)}
	\centering
		\renewcommand{\arraystretch}{1.8}
		\begin{tabular}{c c c c} 
			\hline
			\textbf{Training Data Domains} &  \textbf{SWIR} & \textbf{MWIR} & \textbf{LWIR} \\ 
			\hline
			%CosFace & 67.19 & 1.56 & 1.56 \\
			%\hline
			%ArcFace & 65.62 & 1.56 & 1.56 \\
            VIS only & 38.10 & 9.52 & 11.11 \\
            VIS + SWIR & \textbf{79.37} & 34.92 & 25.40 \\
            VIS + SWIR + MWIR+ LWIR & 77.78 & \textbf{77.78} & \textbf{77.78} \\
			%\hline
			\hline
	\end{tabular}
\end{table}

\textbf{Domain-Aware Sampling.} 
Table \ref{tab:da_sampling} compares the performance of the domain-aware sampler against a random sampler. The results show that when batches are constructed such that each subject has images from all four domains (VIS, SWIR, MWIR, and LWIR), the performance remains more consistent across domains. Notably, while this constraint filters out subjects lacking data from all four domains - potentially reducing the training dataset size - there is still an overall improvement in performance. 

Next, we analyze how domain-aware sampling enhances the triplet loss hard-mining algorithm, leading to improved domain consistency.

\begin{table}[t!]
	\renewcommand{\arraystretch}{1.5}
	\caption{\label{tab:da_sampling}Comparison of sampling strategies - Rank-1 (\%)}
	\centering
		\renewcommand{\arraystretch}{1.8}
		\begin{tabular}{c c c c} 
			\hline
			\textbf{Sampling Strategy} &  \textbf{SWIR} & \textbf{MWIR} & \textbf{LWIR} \\ 
			\hline
			%CosFace & 67.19 & 1.56 & 1.56 \\
			%\hline
			%ArcFace & 65.62 & 1.56 & 1.56 \\
            Random Sampling & \textbf{79.37} & 73.02 & 71.43 \\
            Domain-Aware Sampling & 77.78 & \textbf{77.78} & \textbf{77.78} \\
			%\hline
			\hline
	\end{tabular}
\end{table}

\textbf{Analysis of Triplet-Loss Hard-Mining Statistics.} 
To compute the triplet-loss \cite{hermans2017defense}, we utilize a standard hard mining strategy which selects positive pairs with the greatest intra-class distance and negative pairs with the shortest inter-class distance within each batch. Put simply, a hard positive pair is a pair of features belonging to the same class that are far apart, while a hard negative pair consists of features from different classes that are close together.  

Given that each batch consists of images from four different domains, it is expected that hard positive pairs will predominantly be selected from different domains, while hard negative pairs will mostly originate from the same domain. This occurs because, although cross-entropy loss enforces identity consistency, it does not explicitly enforce domain invariance.

At the beginning of training, 92\% of hard positive pairs were selected from different domains, which slightly decreased to 88\% by the end. Meanwhile, the percentage of hard negative pairs chosen from the same domain dropped significantly from 87\% to 45\% over the course of training.

Cross-entropy loss facilitates clustering of features of the same identity across domains, improving intra-class compactness. However, the continued selection of hard positive pairs from different domains indicates that class clusters retain some domain-specific characteristics, even at the end of training. By the end of training, hard negative pairs are selected almost equally from the same and different domains, indicating that features from separate classes in different domains are converging. This suggests that reducing intra-class inter-domain distance is inherently more challenging than reducing inter-class intra-domain distance.

\textbf{Training Strategies for Smaller Datasets.} 
Table \ref{tab:lora_vs_finetuning} presents results across training datasets of varying sizes, ranging from 8 to 96 subjects. We examine whether LoRA-based \cite{hu2021lora} finetuning offers any advantages over standard finetuning. LoRA (Low-Rank Adaptation) freezes the model parameters and only trains low-rank matrices that approximate the larger weight matrices, significantly reducing the number of trainable parameters. During inference, the learned low-rank decomposition matrices are combined with the original model parameters to produce the final embeddings. We apply LoRA only to the self-attention layers of the ViT-Base model, freezing all the weights. This follows standard practice for efficient fine-tuning while preserving the pretrained feature representations.

Our results indicate that LoRA outperforms standard finetuning across all domains when the training dataset size is very small (ranging from 8 to 32 subjects). However, as the number of subjects increases, standard finetuning gradually overcomes its limitations, ultimately suprassing LoRA.

% LoRA vs finetuning on smaller datasets?
\begin{table}[t]
	\renewcommand{\arraystretch}{1.2}
	\caption{\label{tab:lora_vs_finetuning}Training Strategies with small datasets - Rank-1 (\%)}
	\centering
		\renewcommand{\arraystretch}{1.2}
		\begin{tabular}{c c c c c} 
			\hline
			\textbf{Training Strategy} & \textbf{\#subjects} & \textbf{SWIR} & \textbf{MWIR} & \textbf{LWIR} \\ 
			\hline
			Standard finetuning & 8 & 49.21 & 33.33 & 34.92 \\
			LoRA-based finetuning & 8 & \textbf{55.56} & \textbf{39.68} & \textbf{38.10} \\
			\hline
			Standard finetuning & 16 & 61.90 & \textbf{47.62} & 41.27\\
			LoRA-based finetuning & 16 & \textbf{65.08} & \textbf{47.62} & \textbf{49.21}\\
            \hline
            Standard finetuning & 32 & 61.90 & 60.32 & 49.21\\
			LoRA-based finetuning & 32 & \textbf{66.67} & \textbf{71.43} & \textbf{63.49}\\
            \hline
            Standard finetuning & 48 & 68.25 & \textbf{65.08} & 58.73\\
			LoRA-based finetuning & 48 & \textbf{71.43} & 60.32 & \textbf{65.08}\\
            \hline
            Standard finetuning & 96 & \textbf{82.54} & \textbf{71.43} & \textbf{66.67}\\
			LoRA-based finetuning & 96 & 74.60 & 68.25 & \textbf{66.67}\\
            \hline
	\end{tabular}
\end{table}

\iffalse
\textbf{Comparison with DEEN \cite{zhang2023diverse}:}
Table \ref{tab:baseline_comparison} presents the 1:N body identification results on IJB-MDF, comparing our model with the DEEN model \cite{zhang2023diverse}. We train the DEEN model on IJB-MDF following the same settings provided in their publicly available code. Since DEEN requires paired VIS and IR images as input, we generate randomized image pairs, that incorporate IR images from SWIR, MWIR and LWIR domains. The results indicate that our model significantly outperforms the DEEN model.

\begin{table}[t]
	\renewcommand{\arraystretch}{1.2}
	\caption{\label{tab:baseline_comparison}1:N Body Identification on IJB-MDF - Rank-1 Retrieval Rates (\%)}
	\centering
		\renewcommand{\arraystretch}{1.2}
		\begin{tabular}{c c c c } 
			\hline
			\textbf{Model} &  \textbf{SWIR} & \textbf{MWIR} & \textbf{LWIR} \\ 
			\hline
			DEEN \cite{zhang2023diverse} & 23.81 & 23.81 & 34.92\\
            Our Model & \textbf{77.78} & \textbf{77.78} & \textbf{77.78} \\
			\hline
	\end{tabular}
\end{table}
\fi
% Norm Pooling?
%%%%%%%%%%%%%%%%%%%%%%%%%%%%%%%%%%%%%%%%%%%%%%%%%%%%%%%%%%%%%%%%%%%%%%%%%%%%%%%%
\section{CONCLUSIONS}

In this paper, we demonstrated that body-based embeddings outperform face-embeddings for cross-spectral biometric recognition on the IJB-MDF dataset, particularly in the medium- and long-wave infrared (MWIR \& LWIR) domains. The lower resolution of MWIR and LWIR imagery has a greater impact on face embeddings than on body embeddings, highlighting the robustness of body-based features. By leveraging a VIS-pretrained vision transformer model that fuses local semantic features with global features, we show that simple finetuning is an effective strategy for mitigating domain discrepancies in cross-spectral body recognition. Our experiments reveal that the local semantic features corresponding to the face, torso and lower body have a significantly greater impact in the LWIR domain compared to SWIR, where facial features play a more dominant role. This approach is validated on both the IJB-MDF and LLCM datasets, where we achieve state-of-the-art mAP scores on LLCM. These findings highlight the importance of leveraging pretrained models for challenging cross-spectral biometric recognition tasks, rather than relying solely on modifying standard VIS architectures with complex domain adaptation modules and loss functions. 

%%%%%%%%%%%%%%%%%%%%%%%%%%%%%%%%%%%%%%%%%%%%%%%%%%%%%%%%%%%%%%%%%%%%%%%%%%%%%%%%
\section{ACKNOWLEDGMENTS}
SH and RC are supported by the BRIAR project. This research is based upon work supported in part by the Office of the Director of National Intelligence (ODNI), Intelligence Advanced Research Projects Activity (IARPA), via [2022-21102100005]. The views and conclusions contained herein are those of the authors and should not be interpreted as necessarily representing the official policies, either expressed or implied, of ODNI, IARPA, or the U. S. Government. The US. Government is authorized to reproduce and distribute reprints for governmental purposes
notwithstanding any copyright annotation therein.

%%%%%%%%%%%%%%%%%%%%%%%%%%%%%%%%%%%%%%%%%%%%%%%%%%%%%%%%%%%%%%%%%%%%%%%%%%%%%%%%

%%%%%%%%%%%%%%%%%%%%%%%%%%%%%%%%%%%%%%%%%%%%%%%%%%%%%%%%%%%%%%%%%%%%%%%%%%%%%%%%
\section*{ETHICAL IMPACT STATEMENT}
% Ethical impact of our work:
The dataset used in this paper, IJB-MDF, was collected with the informed consent of all participants. All the sample images used in this paper feature subjects who have explicitly consented to the use of their facial and non-facial information in publications. 

%The primary applications of cross-spectral biometric recognition include security, defense, surveillance, retail, healthcare, transportation, and public safety. 
%While we recognize the potential for misuse by malicious actors that is inherent in any biometric recognition system, we believe that the benefits of this research outweigh the associated risks. Furthermore, infrared sensors are more specialized and less prevalent than visible cameras, making this research less susceptible to misuse.

% Risks - individual harm and negative societal impacts. 
Some ethical risks associated with cross-spectral body identification include privacy violations, misuse by governments and law enforcement, misuse by malicious actors, biometric data security, and discrimination driven by racial or systemic biases. Since thermal imagery can be captured passively without external light sources, it enables surveillance without subjects' awareness. Face and body recognition systems are known to reflect biases arising from both the training data and inherent limitations within the network architecture. Since cross-spectral datasets are typically smaller than VIS-only datasets, the likelihood of biases is higher. 

% Strategies to mititgate risks
Carefully curating large-scale cross-spectral datasets that are demographically balanced can contribute to the development of fairer and less biased cross-spectral biometric systems. However, mitigating potential misuse remains a significant challenge. The most effective approach would be regulating access to this technology, preventing it from falling into the hands of malicious actors. Yet, given the current state of deep learning research, enforcing such restrictions meaningfully is nearly impossible due to the widespread availability of GPUs and the ease of finetuning open-source pretrained models on datasets of interest. Another approach to counteracting potential harm is to enhance public perception of deep learning and AI technologies, fostering greater interest and adoption for societal benefit and the greater good. Reducing fear and misinformation surrounding AI is crucial, as unfounded skepticism could lead to unnecessary suppression of these technologies amond those seeking to use them ethically. Meanwhile, those with malicious intent, who are often more motivated, may continue to exploit AI advancements, resulting in a greater concentration of power in the wrong hands.

% Benefits - rescue operations in challenging conditions of smoke, rain, snow etc. nighttime surveillance, military reconnaissance, 
The capability of cross-spectral biometric systems to operate in adverse environmental conditions such as smoke, fog, rain, or snow makes them invaluable for search and rescue operations during natural disasters like earthquakes, storms, and wildfires. Their ability to function effectively even in low-light or complete darkness, also enhances their utility in surveillance and security, particularly in critical areas such as borders, war zones, and high-density, low-lit environments like concerts. On a smaller scale, these systems can enhance daily life by enabling contactless biometric authentication in low-light or poor visibility conditions, where traditional VIS-based biometric systems often struggle.

%In conclusion, we believe that there is a lot of scope for applying cross-spectral biometric systems for the betterment of society 

% (Since IR is more invariant than visible to clothing changes, and there is higher chance of suspicious individuals trying to evade detection using such tactics...) 

%This year, at FG 2025, we will be introducing a new requirement that authors submit an Ethical Impact Statement as part of the submission process. (Note that this requirement applies to all short and long papers submitted to the main track. Each special track may have its own rules.) To support authors in meeting this requirement, as well as reviewers in assessing it, we have prepared this document with general guidelines, a checklist for authors and reviewers to complete, and answers to some frequently asked questions. As this is a new policy, we welcome questions and feedback as we refine it and develop a shared understanding.
%%%%%%%%%%%%%%%%%%%%%%%%%%%%%%%%%%%%%%%%%%%%%%%%%%%%%%%%%%%%%%%%%%%%%%%%%%%%%%%%

{\small
\bibliographystyle{ieee}
\bibliography{egbib}
}

\end{document}